\documentclass[11pt]{article}

\usepackage{acl}
\usepackage{times}
\usepackage{latexsym}
\usepackage[T1]{fontenc}
\usepackage[utf8]{inputenc}
\usepackage{microtype}
\usepackage{inconsolata}
\usepackage{graphicx}
\usepackage{amssymb} 
\usepackage{array}
\usepackage{booktabs}
\usepackage{multirow}
\usepackage{tabularx} 
\usepackage{threeparttable} 
\usepackage{rotating}  
\usepackage{amsmath}
\usepackage{placeins}
\usepackage{dblfloatfix}

\title{Language Family Matters: Evaluating LLM-Based ASR Across Linguistic Boundaries}

\author{
Yuchen Zhang\textsuperscript{1,2} \and Ravi Shekhar\textsuperscript{1,2} \and Haralambos Mouratidis\textsuperscript{1,2} \\
\textsuperscript{1}Institute for Analytics and Data Science, University of Essex \\
\textsuperscript{2}School of Computer Science and Electronic Engineering, University of Essex \\  
\texttt{\{yuchen.zhang,r.shekhar,h.mouratidis\}@essex.ac.uk}
}

\begin{document}
\maketitle
\begin{abstract}

Large Language Model (LLM)-powered Automatic Speech Recognition (ASR) systems achieve strong performance with limited resources by linking a frozen speech encoder to a pretrained LLM via a lightweight connector. Prior work trains a separate connector per language, overlooking linguistic relatedness. We propose an efficient and novel connector-sharing strategy based on linguistic family membership, enabling one connector per family, and empirically validate its effectiveness across two multilingual LLMs and two real-world corpora spanning curated and crowd-sourced speech. Our results show that family-based connectors reduce parameter count while improving generalization across domains, offering a practical and scalable strategy for multilingual ASR deployment.

\end{abstract}

\section{Introduction}

Automatic speech recognition (ASR) has progressed rapidly with advances in model architectures and training methods. These advances have allowed ASR to evolve from single-language systems to models capable of covering several languages in a unified framework. 
Multilingual ASR systems,
from early multilingual acoustic models \citep{schultz2001language, kamper2021improved, abate2020multilingual} to modern universal encoders such as wav2vec 2.0, HuBERT, XLS-R, and Whisper \citep{baevski2020wav2vec, hsu2021hubert, babu2022xls, radford2023robust}, demonstrate the potential of broad multilingual coverage but also face a continuing challenge in supporting a wide range of languages while preserving efficiency and accuracy.

One way forward is to combine speech encoders with large language models (LLMs) to build SpeechLLMs, which bring together the acoustic representation power of speech encoders and the reasoning and generative capabilities of LLMs \citep{xue2024ideal, verdini2024connect, fan2025alignformer}. Some approaches connect speech encoders and LLMs by training large end-to-end models, such as Qwen-Audio, Salmonn, and decoder-only pipelines \citep{chu2023qwen, tang2023salmonn, wu2023decoder}. However, such end-to-end SpeechLLMs are typically parameter heavy and computationally expensive, making them costly to train and fine-tune across languages or domains. 
A more parameter-efficient alternative is the use of lightweight connectors or adapters, which map acoustic features from speech encoders into the text space of LLM decoders, while the encoder and decoder can be either frozen or trainable depending on task requirements \citep{ma2024embarrassingly, ma2025speech, kumar2025performance, fong2025speech,mundnich2025zero}. 
Despite their progress, current studies on SpeechLLMs mainly focus on architectural design, raising the open question of how the level of data granularity affects multilingual ASR. To address this gap, we focus on two research questions:

RQ1: \textit{Which level of data granularity, individual language or language family, is more effective for multilingual ASR}? 

RQ2: \textit{How well do connectors generalize across domains?}

This paper addresses these gaps with a large-scale study spanning ten language families comprising nearly forty languages, evaluated across two multilingual datasets and two LLM backbones. We systematically evaluate connectors trained at the family and language levels and further examine their robustness under cross-domain transfer. 
Our main contributions are:
(1) We systematically compare language and family connectors for multilingual ASR, covering ten language families comprising nearly forty languages. (2) We conduct a detailed evaluation of cross-domain generalization, assessing how connectors trained on one corpus transfer to a different domain. (3) We provide empirical evidence that family-level connectors strike the best balance between coverage and specificity, resulting in lower WERs and greater stability.

\section{Method}
\vspace{ -1 em}
\begin{figure}[htbp]
    \centering
    \includegraphics[width=0.5\textwidth]{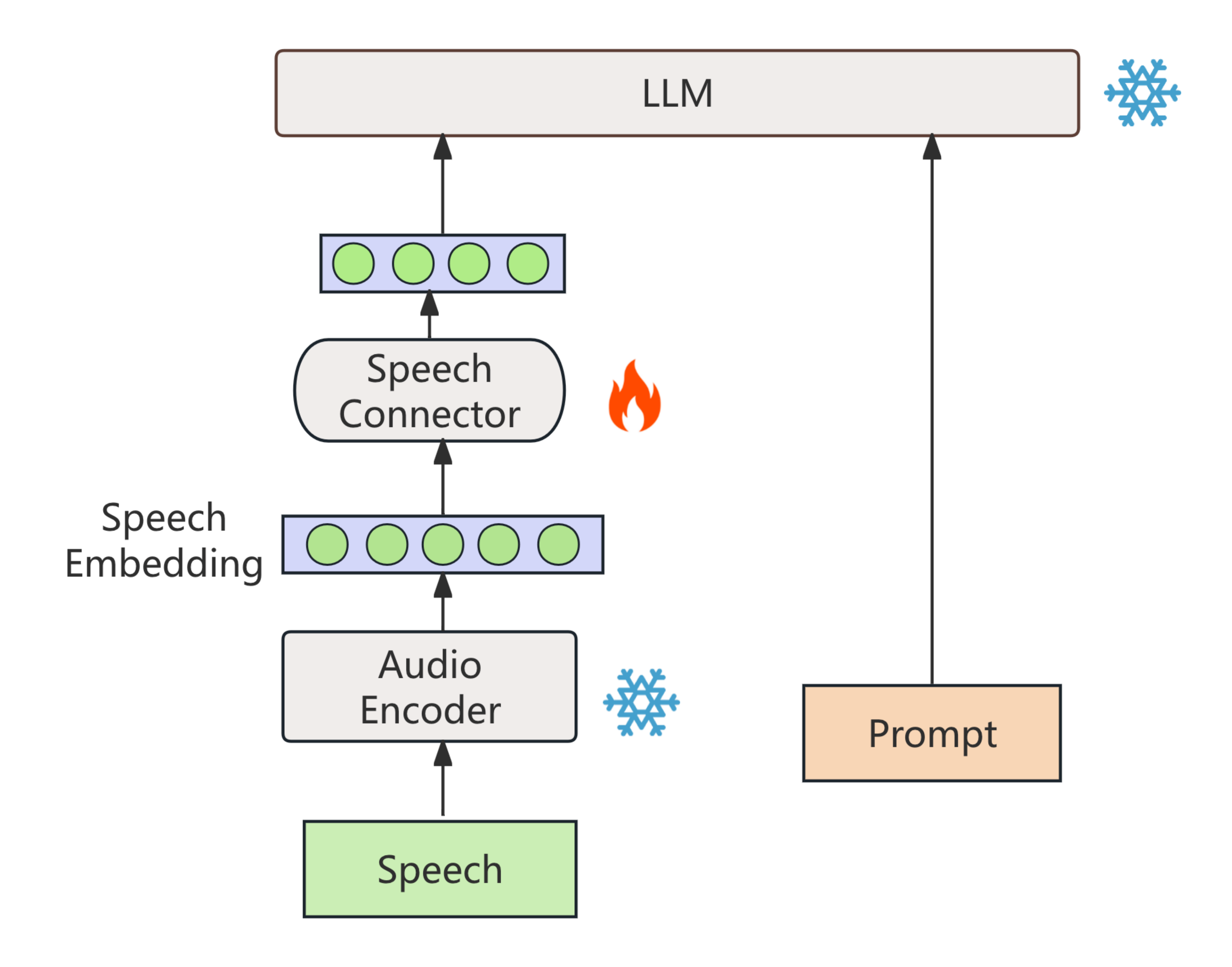}
    \vspace{ -2 em}
    \caption{Overall framework for multilingual ASR.}
    \label{fig:framework}
\end{figure}

We adopt an Encoder–Connector–Decoder architecture for the multilingual ASR, as shown in Figure \ref{fig:framework}. Given an input speech signal $S$, the audio encoder extracts a sequence of high-dimensional acoustic embeddings:

\begin{equation}
    \mathbf{H}_{\text{speech}} = \mathcal{E}(S) \in \mathbb{R}^{B \times T \times E},
\end{equation}

where $\mathcal{E}$ denotes the audio encoder, and $B$, $T$, and $E$ denote the batch size, 
the number of acoustic frames, and the encoder hidden dimension, respectively.

To align the encoder outputs with the embedding space of the LLM, the speech representations $\mathbf{H}_{\text{speech}}$ are first downsampled by a factor $K$, where every $K$ consecutive frames are concatenated into a single vector. The stacked features are then transformed through two successive linear layers:
\begin{align}
\mathbf{H}_{\text{proj}} &= \mathcal{L}_2\!\left( \sigma_{\textsc{gelu}}\!\big(\mathcal{L}_1(\mathbf{H}_{\text{stacked}})\big) \right), \\
\mathbf{H}_{\text{stacked}} &= \mathcal{D}_K(\mathbf{H}_{\text{speech}}) 
\in \mathbb{R}^{B \times \tfrac{T}{K} \times (E \cdot K)}, 
\end{align}
where $\mathcal{D}_K(\cdot)$ denotes the downsampling operator, $\mathcal{L}_1$ and $\mathcal{L}_2$ denote a linear projector, and $\sigma_{\textsc{gelu}}(\cdot)$ is the Gaussian Error Linear Unit (GELU) activation function.

The projected representations $\mathbf{H}_{\text{proj}}$ are then passed to the LLM decoder, which generates the transcript in an autoregressive manner. At each step $t$, the decoder predicts the next token $y_t$ given the previous tokens $y_{1:t-1}$ and the projected speech features $\mathbf{H}_{\text{proj}}$. Decoding terminates once the end-of-sequence token is generated, at which point the complete transcription is obtained. It should be noted that throughout this process, the speech encoder and the LLM decoder remain frozen, while only the connector is trainable.

\section{Experiments and Discussion}

\subsection{Datasets}

\begin{table*}[ht]
\centering
\footnotesize
\resizebox{\linewidth}{!}{
\begin{threeparttable}
\caption{Selected languages and training set statistics in hours (FLEURS / CommonVoice).}
\label{tab:data}
\setlength{\tabcolsep}{0.1pt}
\begin{tabular}{p{0.12\textwidth}*{5}{p{0.19\textwidth}}}
\toprule
\textbf{Family} & \multicolumn{5}{c}{\textbf{Languages (FLEURS / CommonVoice, hours)}} \\
\midrule
Afro-Asiatic  & Amharic (11.1/0.9) & Arabic (6.1/32.4) & Hausa (13.6/2.3) & Hebrew (9.5/1.2) & Maltese (9.9/2.4) \\
Baltic & Latvian (6.5/23.3) & Lithuanian (9.8/11.8) &  &  &  \\
Celtic & Irish (12.1/0.6) & Welsh (12.2/11.5) &  &  &  \\
Dravidian & Malayalam (10.1/1.4) & Tamil (8.7/83.7) & Telugu (7.9/0.1) &  &  \\
Germanic & Danish (7.5/4.2) & Dutch (7.7/54.2) & English (7.5/100) & German (9.0/100) & Swedish (8.4/9.1) \\
Indo-Iranian & Bengali (10.7/34.2) & Hindi (6.7/5.8) & Persian (12.1/31.6) & Punjabi (6.4/1.2) & Urdu (7.0/8.4) \\
Niger-Congo & Igbo (13.8/0.01) & Swahili (13.5/69.9) & Yoruba (10.0/2.4) &  &  \\
Romance & French (10.3/100) & Galician (6.7/96.5) & Portuguese (10.2/26.3) & Romanian (10.1/5.7) & Spanish (8.8/100) \\
Slavic & Belarusian (9.5/100) & Polish (9.2/35.5) & Russian (8.1/38.1) & Serbian (10.7/1.8) & Slovenian (7.8/1.5) \\
Turkic & Azerbaijani (9.3/0.2) & Kazakh (11.8/0.8) & Kyrgyz (9.3/2.3) & Turkish (8.3/43.8) &  \\
\bottomrule
\end{tabular}
\end{threeparttable}}
\end{table*}

We adopt two widely used multilingual speech corpora for our experiments, FLEURS \citep{conneau2023fleurs} and CommonVoice\_22 \citep{ardila2020common} \footnote{For simplicity, we will use CommonVoice to refer to CommonVoice\_22 in the rest of the paper.}. To ensure diversity while maintaining balance across language families, we focus on seven major families, including Afro-Asiatic, Austronesian, Dravidian, Indo-European, Niger-Congo, Turkic, and Uralic, as shown in Table \ref{tab:data}. Since each family contains a large number of languages, we sample up to five representatives per family. For Indo-European, which is especially diverse, we further group languages by branch. The final set of languages used in our study is summarized in Table~\ref{tab:data}, where we report the available training hours from both FLEURS and CommonVoice for each selected language. We cap the training data at 100 hours per language, as some CommonVoice languages contain extremely large training sets.

\subsection{Experimental Setting}

In this study, we employ Whisper-large-v3 \cite{radford2022whisper} as the speech encoder and adopt Gemma-2-2b \cite{gemma_2024}, and Salamandra-2b \cite{gonzalez2025salamandra} as LLM decoders \footnote{For simplicity, we will use Gemma and Salamandra to refer to Gemma-2-2b and Salamandra-2b in the rest of the paper.}. Two types of connectors are evaluated: a language-specific connector \textsc{LangConn}, trained on data from a single language, and a family-level connector \textsc{FamConn}, trained on the combined data of all languages within a given family. In all of our experiments, the parameters of both the speech encoder and the LLM are frozen, and only the connector is trainable. Every connector was trained for 10 epochs with early stopping, using a batch size of 10, the AdamW optimizer, a learning rate of 1e-4, and a weight decay of 1e-6. During inference, beam search is used for decoding, with the beam size set to 2. The prompt is fixed for all experiments as “Transcribe the speech to text:”. Training and evaluation are conducted on four NVIDIA A10 GPUs, and performance is assessed using the Word Error Rate (WER) metric.

\begin{table*}[!ht]
\centering
\footnotesize
\resizebox{\linewidth}{!}{
\begin{threeparttable}
\caption{ Family level WER$\downarrow$ (\%) across datasets and connector types. 
$\Delta$ = Fam -- Lang. Best in \textbf{bold}.}
\label{tab:rq1_wer}
\begin{tabular}{lllllll|llllll}
\toprule
\multirow{3}{*}{\textbf{Family}} 
& \multicolumn{6}{c}{\textbf{FLEURS}} 
& \multicolumn{6}{c}{\textbf{CommonVoice}} \\
\cmidrule(lr){2-7} \cmidrule(lr){8-13}
& \multicolumn{3}{c}{\textbf{Salamandra}} & \multicolumn{3}{c}{\textbf{Gemma}}
& \multicolumn{3}{c}{\textbf{Salamandra}} & \multicolumn{3}{c}{\textbf{Gemma}} \\
\cmidrule(lr){2-4} \cmidrule(lr){5-7} \cmidrule(lr){8-10} \cmidrule(lr){11-13}
& \textbf{Lang} & \textbf{Fam} & $\Delta$ 
& \textbf{Lang} & \textbf{Fam} & $\Delta$
& \textbf{Lang} & \textbf{Fam} & $\Delta$
& \textbf{Lang} & \textbf{Fam} & $\Delta$ \\
\midrule

Afro-Asiatic 
& \textbf{70.56} & 92.69 & +22.13  
& \textbf{44.77} & 46.66 & +1.89 
& \textbf{93.90} & 109.56 & +15.66  
& 105.89 & \textbf{91.60} & -14.29 \\

Baltic       
& 113.21 & \textbf{39.64} & -73.57 
& 31.23  & \textbf{27.56} & -3.67  
& 123.26 & \textbf{91.75} & -31.51 
& 61.63  & \textbf{59.96} & -1.67 \\

Celtic       
& 116.78 & \textbf{113.73} & -3.05 
& 53.58  & \textbf{53.11}  & -0.47 
& 211.35 & \textbf{177.34} & -34.01  
& 119.70 & \textbf{92.65}  & -27.05 \\

Dravidian    
& \textbf{28.34} & 38.60 & +10.26  
& \textbf{27.22} & 28.34 & +1.12  
& 125.18 & \textbf{71.47} & -53.71 
& 105.45 & \textbf{85.59} & -19.86 \\

Germanic
& 23.37 & \textbf{15.67} & -7.70
& 17.26 & \textbf{14.22} & -3.04
& 77.71 & \textbf{33.55} & -44.16
& 44.86 & \textbf{25.85} & -19.01 \\

Indo-Iranian 
& 49.20 & \textbf{46.09} & -3.11   
& 39.43 & \textbf{29.60} & -9.83  
& 73.39 & \textbf{67.87} & -5.52 
& 101.68 & \textbf{97.14} & -4.54 \\

Niger-Congo  
& 53.59 & \textbf{52.39} & -1.20   
& \textbf{46.25} & 48.14 & +1.89 
& 278.72 & \textbf{85.30} & -193.42   
& \textbf{82.43} & 90.48 & +8.05 \\

Romance
& 37.47 & \textbf{11.15} & -26.32
& 12.80 & \textbf{10.58} & -2.22
& 62.18 & \textbf{25.49} & -36.69
& 38.42 & \textbf{29.53} & -8.89 \\

Slavic       
& 95.96 & \textbf{24.61} & -71.35   
& 25.07 & \textbf{21.69} & -3.38  
& 180.62 & \textbf{44.69} & -135.93  
& 47.54 & \textbf{41.58} & -5.96 \\

Turkic       
& 60.09 & \textbf{45.45} & -14.64  
& 26.29 & \textbf{24.55} & -1.74  
& 82.02 & \textbf{69.07} & -12.95 
& 79.16 & \textbf{71.59} & -7.57 \\

\bottomrule
\end{tabular}
\end{threeparttable}}
\end{table*}

\subsection{Results and Discussion}

\begin{table}[ht]
\centering
\scriptsize
\setlength{\tabcolsep}{3pt} 
\begin{threeparttable}
\caption{Family level cross-domain WER$\downarrow$ (\%) using Gemma. $\Delta$ = Fam -- Lang. Best in \textbf{bold}.}
\label{tab:rq2_fam}
\begin{tabular}{l|lll|lll}
\toprule
\multirow{2}{*}{\textbf{Family}} 
& \multicolumn{3}{c|}{\textbf{CV $\rightarrow$ FL}} 
& \multicolumn{3}{c}{\textbf{FL $\rightarrow$ CV}} \\
\cmidrule(lr){2-4} \cmidrule(lr){5-7}
& \textbf{Lang} & \textbf{Fam} & $\Delta$
& \textbf{Lang} & \textbf{Fam} & $\Delta$ \\
\midrule
Afro-Asiatic & 154.16 & \textbf{150.60} & -3.56  & 79.64 & \textbf{67.89} & -11.75 \\
Baltic       & \textbf{111.54} & 122.68 & +11.14 & 73.84 & \textbf{49.66} & -24.18 \\
Celtic       & 179.45 & \textbf{169.31} & -10.14 & \textbf{104.15} & 105.71 & +1.56 \\
Dravidian    & \textbf{128.39} & 150.83 & +22.44 & \textbf{75.81} & 76.23 & +0.42 \\
Germanic     & 124.01 & \textbf{56.03} & -67.98 & 26.41 & \textbf{23.66} & -2.75 \\
Indo-Iranian & 144.19 & \textbf{132.79} & -11.40  & 90.92 & \textbf{84.98} & -5.94 \\
Niger-Congo  & \textbf{127.46} & 132.63 & +5.17  & \textbf{69.03} & 89.46 & +20.43 \\
Romance      & 100.25 & \textbf{72.91} & -27.34 & 27.51 & \textbf{26.23} & -1.28 \\
Slavic       & 110.59 & \textbf{102.64} & -7.95  & 103.42 & \textbf{65.81} & -37.61 \\
Turkic       & \textbf{118.94} & 130.80 & +11.86 & 47.76 & \textbf{46.03} & -1.73 \\
\bottomrule
\end{tabular}
\end{threeparttable}
\end{table}

\paragraph{RQ1: Impact of Data Granularity on Multilingual ASR}\mbox{}\\
To investigate whether modeling linguistic structure at the family level benefits multilingual ASR, we compare the performance of \textsc{LangConn} and \textsc{FamConn} across both LLMs and datasets. Table~\ref{tab:rq1_wer} summarizes the WER at the family level, while Table~\ref{tab:rq1_detailed} (in Appendix \ref{sec:appendix}) provides detailed language level WERs. It should be noted that for the \textsc{FamConn}, the family-level evaluation is performed directly on the merged test set containing all languages within a family. For the \textsc{LangConn}, predictions and references are obtained separately for each language, then concatenated before calculating family-level WER.

From the results, we can see that across all families and configurations, \textsc{FamConn} outperforms \textsc{LangConn} in almost all cases.
For instance, on Salamandra with FLEURS, \textsc{FamConn} achieves 15.67\% WER for Germanic and 11.15\%  for Romance, compared to 23.37\%  and 37.47\%  with \textsc{LangConn}. The gap widens on CommonVoice, with Germanic improving from 77.71\%  to 33.55\%, and Romance from 62.18\%  to 25.49\%. 
Gains are particularly pronounced in high-variance families such as Slavic, Baltic, and Romance, where shared morphological and phonological patterns may benefit from joint modeling. For instance, Belarusian and Latvian show WER reductions exceeding 70\% on FLEURS, indicating strong transferability within families.

These benefits are robust across both Gemma and Salamandra, and across datasets. On CommonVoice, which is crowd-sourced and more acoustically diverse, the relative advantage of family-based connectors is even more pronounced, suggesting improved generalization under domain shift. For example, the Germanic family shows a reduction from 77.71\% to 33.55\% WER with Salamandra on CommonVoice. Additionally, while \textsc{FamConn} outperforms \textsc{LangConn} for both Gemma and Salamandra in most cases, the magnitude of improvement differs. Salamandra shows greater instability in language-specific transcription, including repetition and language drift. In these cases, family-level grouping provides substantial corrective effects, leading to larger relative gains for \textsc{FamConn}. Gemma exhibits more stable performance, resulting in smaller but more consistent gains across languages. These trends indicate that the benefit of family-level sharing interacts with the inherent stability of the underlying LLM backbone.

However, \textsc{FamConn} are not universally superior. In a minority of cases, such as the Afro-Asiatic and Dravidian families, \textsc{FamConn} underperforms.
The families in which FAMCONN performs worse than LANGCONN tend to be those where historical family membership does not align well with the acoustic or phonological similarity required for stable connector sharing. For example, in the Afro-Asiatic family, Arabic, Hebrew, Amharic, and Maltese differ substantially in script and phonological structure \citep{fabri2014linguistic}. Similarly, the Dravidian family is internally diverse, with clear cross-language variation in phonology \citep{kolipakam2018bayesian}. These factors may limit the extent to which a single family-level connector can generalise across all languages in the group. By contrast, families such as Germanic, Romance, Slavic, and Baltic display more consistent improvements under \textsc{FamConn}, suggesting that family-level sharing is more effective when intra-family conditions are relatively coherent. 

Overall, our results indicate that the benefits of family-level pooling depend on the interaction between linguistic similarity and dataset quality, and that genealogical relatedness alone does not guarantee improved performance. These observations also suggest that heterogeneous families may benefit from finer-grained alternatives such as sub-family connectors or lightweight language-specific adapters.

\paragraph{RQ2: Cross-Domain Generalization ability of Connectors} \mbox{}\\
Table \ref{tab:rq2_fam} presents the family-level WERs across language families when training and evaluation occur on mismatched speech domains, i.e., connectors were trained on FLEURS but tested on CommonVoice, and vice versa. The results compare \textsc{FamConn} and \textsc{LangConn} for both directions of transfer. Detailed language-specific WERs are summarized in Table \ref{tab:rq2_crossdomain} available in Appendix \ref{sec:appendix}.

When training on CommonVoice and evaluating on FLEURS, \textsc{FamConn} outperforms \textsc{LangConn} in more cases. For example, in the Germanic family, the WER drops from 124.01\% using \textsc{LangConn} to 56.03\% using \textsc{FamConn}, representing a substantial reduction in error. Similar trends are observed in the Slavic and Romance families, where \textsc{FamConn} achieves better generalization. However, we also observe that for some families, such as Dravidian, \textsc{LangConn} surpasses \textsc{FamConn} over 20\%. Overall, under this setting, \textsc{FamConn} can yield large gains in certain families, while \textsc{LangConn} remains competitive and occasionally better for others. 
In the reverse setting, where training is conducted on FLEURS and evaluation takes place on CommonVoice, the benefits of \textsc{FamConn} are clearer. 
\textsc{FamConn} is better across seven out of ten families. 
While the improvements are generally smaller than those observed in the Germanic family in the CV-to-FL setting, we still observe notable gains in some families. For instance, in the Slavic family, the WER is reduced from 103.43\% to 64.81\%, yielding a 37.61\% absolute improvement. Moreover, in this setting, when \textsc{FamConn} underperforms \textsc{LangConn}, the performance gap is typically modest. 
Despite some family-specific exceptions and varying effect sizes across transfer directions, \textsc{FamConn} generally outperforms \textsc{LangConn}. This pattern suggests that family-level representations capture broader phonological and prosodic regularities that support cross-domain transfer.

At the language level, the above pattern becomes even more apparent (see Table \ref{tab:rq2_crossdomain} in Appendix \ref{sec:appendix}). A majority of languages benefit significantly from \textsc{FamConn}, with large reductions in WER. 
For instance, Serbian exhibits a large absolute improvement in both transfer directions, indicating that family-level information provides strong inductive bias for cross-domain generalization. Similarly, Slovenian consistently shows lower error rates under \textsc{FamConn}, suggesting that shared phonological and morphological characteristics within the Slavic family are effectively exploited by family-level connectors.
In contrast, some languages show a strong preference for \textsc{LangConn}.
In these cases, \textsc{FamConn} leads to noticeable performance degradation, with WER increases exceeding 20\%. This behavior can be attributed to the high internal diversity within their respective language families. For instance, for Telugu in the Dravidian family and Punjabi in the Indo-Iranian family, the substantial phonetic and prosodic differences across family members limits the usefulness of family-level representations, making language-specific connectors more effective. \citep{jairam2024few}. 

In summary, our cross-domain evaluation demonstrates that \textsc{FamConn} provides a robust inductive bias for transfer across mismatched speech domains. Specifically, \textsc{FamConn} outperforms \textsc{LangConn} for most language families and individual languages, particularly those with strong intra-family phonological similarity (e.g., Slavic and Germanic). 
We also observe that the effectiveness of \textsc{FamConn} is limited when the target domain introduces greater acoustic and lexical variability. This highlights the importance of accounting for both source- and target-domain characteristics when designing multilingual generalization strategies.

\begin{table}[!h]
\centering
\caption{Family level WER$\downarrow$ (\%) using Gemma across \textsc{LangConn}, \textsc{FamConn} and \textsc{UniConn}. Best in \textbf{bold}. }
\begin{tabular}{lccc}
\toprule
\multirow{2}{*}{Family} & \multicolumn{3}{c}{FLEURS}              \\ \cline{2-4}  & Lang           & Fam            & Uni  
\\ \midrule
Afro-Asiatic & \textbf{44.77} & 46.66 & 54.97 \\
Baltic       & 31.23 & \textbf{27.56} & 29.86 \\
Celtic       & 53.58 & \textbf{53.11} & 63.30 \\
Dravidian    & \textbf{27.22} & 28.34 & 39.36 \\
Germanic     & 17.26 & \textbf{14.22} & 15.37 \\
Indo-Iranian & 39.43 & \textbf{29.60} & 50.38 \\
Niger-Congo  & \textbf{46.25} & 48.14 & 60.96 \\
Romance      & 12.80 & \textbf{10.58} & 11.55 \\
Slavic       & 25.07 & \textbf{21.69} & 22.68 \\
Turkic       & 26.29 & \textbf{24.55} & 29.43 \\
\bottomrule
\end{tabular}
\label{tab:uni_family}
\end{table}

\paragraph{Impact of training data volume} \mbox{}\\
In our experiments, \textsc{LangConn} is trained on data from a single language, whereas \textsc{FamConn} is trained on pooled data from multiple languages within the same family. To validate that the improvements brought by \textsc{FamConn} are tied to linguistic relatedness rather than simply to a larger training data volume, we further train a universal connector, \textsc{UniConn}, on all FLEURS languages pooled together using Gemma, as a complement to \textsc{LangConn} and \textsc{FamConn}.

Table~\ref{tab:uni_family} shows the family-level WER across these three connectors,  while Table~\ref{tab:uni_detailed} (in Appendix \ref{sec:appendix}) provides detailed language level WERs. 
The results show that, across all families, \textsc{FamConn} consistently achieves lower WER than \textsc{UniConn}, even though \textsc{UniConn} is trained with the largest volume of data. At the per-language level, \textsc{UniConn} generally falls between \textsc{LangConn} and \textsc{FamConn} and rarely surpasses the family connector. These results suggest that linguistic relatedness, rather than data volume, is the key factor in the gains observed with \textsc{FamConn}.

\paragraph{Extreme WERs} \mbox{} \vspace{-1 em}\\

We also conduct an analysis of cases with extreme WERs. After manually reviewing the predicted transcripts, we observe a clear distinction between low WER and high WER cases. The extreme WERs mainly arise from repetition and overlong outputs. For instance, for \textsc{LangConn} on Polish in FLEURS with Salamandra, 94.08\% of predictions contain at least one word or phrase repeated three or more times consecutively, and 45\% of predictions have output lengths exceeding 1.25 times the reference length. For more detailed analysis results, please see Table~\ref{tab:extreme} in Appendix \ref{sec:appendix}.

\section{Conclusion}

This study investigated grouping strategies for multilingual ASR, comparing family-level and language-specific connectors across two LLMs, two real-world multilingual datasets, and ten language families covering nearly forty languages. 
The findings demonstrate that family-level grouping consistently achieves lower and more stable WERs than language-specific alternatives. Although language-specific connectors occasionally deliver marginal improvements for individual languages, these gains are outweighed by frequent and sometimes catastrophic failures. Cross-domain testing further highlights that family-level sharing is particularly beneficial for multilingual ASR, where transfer across related languages enhances robustness. Overall, family-level grouping emerges as the more effective and reliable solution for multilingual ASR, striking a balance between transferability and specialization while avoiding the instability inherent in language-specific approaches.

\section{Limitations and Ethical Statement}

\paragraph{Limitations:} Our work investigates speech connectors with two LLM backbones, Gemma-2\_2b and Salamandra\_2b, evaluated on the FLEURS and CommonVoice datasets covering nearly forty languages across ten language families. We have carefully limited our analysis to a maximum of five languages per family to ensure a manageable yet diverse experimental setup, enabling systematic comparisons without compromising generalizability. However, we acknowledge that while this choice allows us to control the grouping size and analyse a broad set of languages, it does not capture other potentially relevant dimensions such as script, morphological type, or subfamily structure. Future work could therefore consider typology-based or script-based groupings, or branch-level subdivisions within large families, to better understand how different forms of cross-lingual similarity influence connector design.

\paragraph{Ethical Statement:} In this work, we have used publicly available models, architectures, and datasets and have not collected any sensitive/private data. The ultimate goal of our study is to contribute to analyzing the effect of language family on  LLM-based ASR to improve multi-lingual ASR that is robust to domain shift.  
research

\section{Acknowledgments}
This work has been partially funded by the European Union’s Horizon Europe research and innovation programme under the project ELOQUENCE (Grant Agreement No. 101135916) and the UKRI. Views and opinions expressed are, however, those of the author(s) only and do not necessarily reflect those of the European Union or the Research Executive Agency. Neither the European Union nor the granting authority can be held responsible for them.

\bibliography{ref}

\clearpage
\appendix
\section{Appendix}
\label{sec:appendix}

In this appendix, we present detailed statistics of the training data and language-level experimental results.

Table \ref{tab:full_splits} reports the total duration (in hours) of the training, validation, and test splits for each language in the FLEURS and CommonVoice datasets used in our experiments. It should be noted that for CommonVoice, the training data are capped at 100 hours per language.

Table \ref{tab:rq1_detailed} presents the full language-level WER results for RQ1, comparing family-level and language-specific connectors across both FLEURS and CommonVoice with two different LLM decoders.

Table \ref{tab:rq2_crossdomain} reports WER results for cross-domain evaluation using Gemma, where models are trained on one dataset and tested on the other.

Table \ref{tab:uni_detailed} provides detailed language-level WER results using Gemma with three different types of connectors, \textsc{LangConn}, \textsc{FamConn} and \textsc{UniConn}, to analyse impact of the training data volume.

Finally, to analyse error patterns in extreme cases, we select three low-WER and three high-WER examples and compute the repetition rate and overlong rate from the predicted transcriptions. Table \ref{tab:extreme} reports these metrics for the selected cases.

\clearpage

\begin{table*}[t]
\centering
\scriptsize
\begin{threeparttable}
\caption{Full dataset statistics in hours (FLEURS and CommonVoice). 
Train/validation/test splits are reported; CommonVoice values are capped at 100 hours.} \label{tab:full_splits}
\begin{tabular}{llcccccc}
\toprule
\multirow{2}{*}{\textbf{Family (Subgroup)}} & \multirow{2}{*}{\textbf{Language}} 
& \multicolumn{3}{c}{\textbf{FLEURS}} & \multicolumn{3}{c}{\textbf{CommonVoice}} \\
\cmidrule(lr){3-5} \cmidrule(lr){6-8}
& & Train & Val & Test & Train & Val & Test \\
\midrule
Afro-Asiatic & Amharic  & 11.1 & 0.65 & 1.61 & 0.9  & 0.39 & 0.44 \\
             & Arabic   & 6.1  & 0.87 & 1.3  & 32.4 & 12.99 & 12.66 \\
             & Hausa    & 13.6 & 1.53 & 3.34 & 2.3  & 0.74 & 0.97 \\
             & Hebrew   & 9.5  & 0.82 & 2.05 & 1.2  & 0.88 & 0.58 \\
             & Maltese  & 9.9  & 1.49 & 3.55 & 2.4  & 2.1  & 2.35 \\
\midrule
Dravidian    & Malayalam & 10.1 & 1.68 & 3.91 & 1.39 & 1.02 & 1.03 \\
             & Tamil     & 8.7  & 1.25 & 2.13 & 83.7 & 19.3 & 19.4 \\
             & Telugu    & 7.9  & 0.89 & 1.45 & 0.08 & 0.08 & 0.08 \\
\midrule
Germanic (Indo-Euro) & Danish  & 7.5 & 1.17 & 2.94 & 4.2  & 3.42 & 3.56 \\
                     & Dutch   & 7.7 & 0.45 & 0.97 & 54.2 & 16.0 & 16.4 \\
                     & English & 7.5 & 1.05 & 1.77 & 100  & 27.3 & 27.1 \\
                     & German  & 9.0 & 1.26 & 3.15 & 100  & 27.5 & 27.6 \\
                     & Swedish & 8.4 & 0.98 & 2.33 & 9.1  & 6.1  & 6.9 \\
\midrule
Romance (Indo-Euro) & French     & 10.3 & 0.8  & 2.0  & 100  & 26.1 & 26.3 \\
                    & Galician   & 6.7  & 1.05 & 2.6  & 96.5 & 18.9 & 19.8 \\
                    & Portuguese & 10.2 & 1.29 & 3.24 & 26.3 & 12.0 & 12.9 \\
                    & Romanian   & 10.1 & 1.08 & 2.53 & 5.7  & 4.2  & 4.6 \\
                    & Spanish    & 8.8  & 1.35 & 3.09 & 100  & 26.8 & 27.0 \\
\midrule
Slavic (Indo-Euro) & Belarusian & 9.5  & 1.65 & 4.03 & 100  & 25.2 & 25.6 \\
                   & Polish     & 9.2  & 0.84 & 2.05 & 35.5 & 14.2 & 14.1 \\
                   & Russian    & 8.1  & 1.08 & 2.5  & 38.1 & 15.5 & 16.0 \\
                   & Serbian    & 10.7 & 0.83 & 2.12 & 1.8  & 1.6  & 1.9 \\
                   & Slovenian  & 7.8  & 0.89 & 2.27 & 1.5  & 1.5  & 1.7 \\
\midrule
Indo-Iranian (Indo-Euro) & Bengali & 10.7 & 1.45 & 3.43 & 34.2 & 16.4 & 16.7 \\
                         & Hindi   & 6.7  & 0.71 & 1.34 & 5.8  & 3.6  & 4.7 \\
                         & Persian & 12.1 & 1.53 & 3.7  & 31.6 & 12.6 & 14.7 \\
                         & Punjabi & 6.4  & 0.75 & 1.84 & 1.2  & 0.49 & 0.74 \\
                         & Urdu    & 7.0  & 0.76 & 0.81 & 8.4  & 6.3  & 6.5 \\
\midrule
Baltic (Indo-Euro) & Latvian    & 6.5  & 1.12 & 2.84 & 23.3 & 12.3 & 12.3 \\
                   & Lithuanian & 9.8  & 1.17 & 2.97 & 11.8 & 7.1  & 7.8 \\
\midrule
Celtic (Indo-Euro) & Irish & 12.1 & 1.49 & 3.46 & 0.6  & 0.58 & 0.64 \\
                   & Welsh & 12.2 & 1.8  & 4.28 & 11.5 & 8.3  & 8.3 \\
\midrule
Niger-Congo & Igbo    & 13.8 & 1.9  & 4.81 & 0.01 & 0.0  & 0.01 \\
            & Swahili & 13.5 & 0.8  & 1.93 & 69.9 & 19.2 & 19.0 \\
            & Yoruba  & 10.0 & 1.71 & 3.77 & 2.4  & 1.35 & 1.91 \\
\midrule
Turkic & Azerbaijani & 9.3 & 1.35 & 3.24 & 0.23 & 0.1  & 0.16 \\
       & Kazakh      & 11.8 & 1.53 & 3.83 & 0.81 & 0.67 & 0.75 \\
       & Kyrgyz      & 9.3 & 1.34 & 3.25 & 2.3  & 2.1  & 2.2 \\
       & Turkish     & 8.3 & 1.12 & 2.61 & 43.8 & 12.1 & 14.1 \\
\bottomrule
\end{tabular}
\end{threeparttable}
\end{table*}

\begin{table*}[ht]
\centering
\scriptsize
\begin{threeparttable}
\caption{Detailed WER$\downarrow$ (\%) per language on FLEURS and CommonVoice datasets with both LLMs. 
$\Delta_f$ = Fam -- Lang. Best in \textbf{bold}}
\label{tab:rq1_detailed}
\begin{tabular}{ll|ll|ll|ll|ll}
\toprule
\multirow{3}{*}{\textbf{Family}} & \multirow{3}{*}{\textbf{Language}} 
& \multicolumn{4}{c|}{\textbf{FLEURS}} & \multicolumn{4}{c}{\textbf{CommonVoice}} \\
\cmidrule(lr){3-6} \cmidrule(lr){7-10}
& & \multicolumn{2}{c}{\textbf{Salamandra\_2b}} & \multicolumn{2}{c|}{\textbf{Gemma-2\_2b}}
  & \multicolumn{2}{c}{\textbf{Salamandra\_2b}} & \multicolumn{2}{c}{\textbf{Gemma-2\_2b}} \\
\cmidrule(lr){3-4} \cmidrule(lr){5-6} \cmidrule(lr){7-8} \cmidrule(lr){9-10}
& & \textbf{Lang} & \textbf{Fam ($\Delta_f$)} 
  & \textbf{Lang} & \textbf{Fam ($\Delta_f$)}
  & \textbf{Lang} & \textbf{Fam ($\Delta_f$)} 
  & \textbf{Lang} & \textbf{Fam ($\Delta_f$)} \\
\midrule
Afro-Asiatic & Amharic & \textbf{97.45} & 98.74 (+1.29)  & \textbf{52.29} & 52.89 (+0.60)  
                         & \textbf{111.53} & 130.68 (+19.15) & \textbf{137.32} & 150.75 (+13.43) \\
             & Arabic  & \textbf{38.95} & 53.66 (+14.71) & \textbf{29.83} & 30.17 (+0.34) 
                         & \textbf{50.69} & 57.65 (+6.96) & \textbf{60.81} & 64.34 (+3.53) \\
             & Hausa   & \textbf{55.72} & 101.22 (+45.50) & \textbf{51.32} & 57.48 (+6.16)  
                         & \textbf{97.57} & 110.61 (+13.04) & 108.88 & \textbf{98.80} (-10.08) \\
             & Hebrew  & \textbf{54.58} & 71.25 (+16.67) & \textbf{42.90} & 46.59 (+3.69) 
                         & \textbf{52.77} & 53.22 (+0.45) & 77.62 & \textbf{65.35} (-12.27) \\
             & Maltese & \textbf{107.49} & 111.56 (+4.07) & \textbf{40.37} & 42.41 (+2.04)  
                         & \textbf{131.25} & 169.01 (+37.76) & 116.29 & \textbf{92.25} (-24.04) \\
\midrule
Baltic  & Latvian    & 115.47 & \textbf{36.30} (-79.17) & 28.27 & \textbf{24.63} (-3.64) 
                         & 137.97 & \textbf{77.39} (-60.58) & 56.64 & \textbf{45.98} (-10.66) \\
        & Lithuanian & 110.76 & \textbf{42.29} (-68.47) & 34.56 & \textbf{30.18} (-4.38) 
                         & 110.80 & \textbf{103.10} (-7.70) & \textbf{59.01} & 70.52 (+11.51) \\
\midrule
Celtic  & Irish  & 122.39 & \textbf{117.09} (-5.30) & 71.54 & \textbf{70.99} (-0.55) 
                         & 302.18 & \textbf{219.56} (-82.62) & 121.59 & \textbf{107.64} (-13.95) \\
        & Welsh  & 115.08 & \textbf{110.97} (-4.11) & 38.74 & \textbf{38.51} (-0.23) 
                         & \textbf{125.94} & 159.53 (+33.59) & 119.93 & \textbf{84.89} (-35.04) \\
\midrule
Dravidian & Malayalam & \textbf{23.65} & 44.79 (+21.14) & \textbf{21.85} & 22.95 (+1.10) 
                         & 112.78 & \textbf{84.91} (-27.87) & \textbf{71.72} & 96.59 (+24.87) \\
          & Tamil     & 33.93 & \textbf{33.63} (-0.30)  & 32.24 & \textbf{31.92} (-0.32) 
                         & 124.42 & \textbf{54.11} (-70.31) & 49.67 & \textbf{68.50} (-18.83) \\ 
          & Telugu    & 35.45 & \textbf{34.60} (-0.85)  & \textbf{34.34} & 35.16 (+0.82) 
                         & 140.21 & \textbf{111.64} (-28.57) & 285.65 & \textbf{176.63} (-109.02) \\
\midrule
Germanic & Danish   & 32.12 & \textbf{20.51} (-11.61) & 24.33 & \textbf{19.24} (-5.09) 
                         & 69.38 & \textbf{43.28} (-26.10) & 44.62 & \textbf{33.42} (-11.20) \\
         & Dutch    & 15.76 & \textbf{13.84} (-1.92)  & 15.33 & \textbf{12.62} (-2.71)  
                         & 104.41 & \textbf{30.24} (-74.17) & 59.33 & \textbf{21.20} (-38.13) \\
         & English  & 13.06 & \textbf{9.37} (-3.69)   & 8.59  & \textbf{6.79} (-1.80) 
                         & 53.96 & \textbf{32.58} (-21.38) & 37.12 & \textbf{23.18} (-13.94) \\
         & German   & 22.56 & \textbf{13.62} (-8.94)  & 17.95 & \textbf{13.34} (-4.61) 
                         & 97.67 & \textbf{27.62} (-70.05) & 37.12 & \textbf{24.75} (-12.37) \\
         & Swedish  & 23.11 & \textbf{18.75} (-4.36)  & 20.61 & \textbf{16.56} (-4.05) 
                         & 46.58 & \textbf{33.93} (-12.65) & 39.05 & \textbf{27.61} (-11.44) \\
\midrule
Indo-Iranian & Bengali & \textbf{41.47} & 52.17 (+10.7)  & 23.95 & \textbf{23.64} (-0.31) 
                         & \textbf{62.68} & 101.19 (+38.51)  & \textbf{101.20} & 101.65 (+0.45) \\
             & Hindi   & 47.53 & \textbf{36.92} (-10.61) & 21.45 & \textbf{19.57} (-1.88) 
                         & 64.67 & \textbf{35.35} (-29.32)  & 72.20 & \textbf{42.99} (-29.21) \\
             & Persian & \textbf{37.56} & 46.01 (+8.45)  & \textbf{25.12} & 30.88 (+5.76) 
                         & \textbf{66.85} & 78.97 (+12.12)  & \textbf{138.66} & 154.45 (+15.79) \\
             & Punjabi & 46.85 & \textbf{45.67} (-1.18)  & 26.71 & \textbf{25.34} (-1.37) 
                         & 83.76 & \textbf{64.62} (-19.14)  & 115.23 & \textbf{48.21} (-67.02) \\
             & Urdu    & 81.18 & \textbf{54.46} (-26.72) & 85.45 & \textbf{33.71} (-51.74) 
                         & 71.21 & \textbf{49.51} (-21.70)  & 69.62 & \textbf{64.44} (-5.18) \\
\midrule
Niger-Congo & Igbo    & 61.72 & \textbf{60.55} (-1.17) & \textbf{55.25} & 56.69 (+1.44) 
                         & \textbf{100.00} & 120.13 (+20.13) & \textbf{100.00} & 108.51 (+8.51) \\
            & Swahili & \textbf{28.87} & 31.63 (+2.76) & \textbf{24.91} & 28.53 (+3.62) 
                         & 482.60 & \textbf{81.12} (-401.48) & 82.16 & \textbf{78.86} (-3.30) \\
            & Yoruba  & 52.24 & \textbf{52.07} (-0.17) & \textbf{45.51} & 46.63 (+1.12) 
                         & \textbf{88.30} & 90.50 (+2.20) & \textbf{75.90} & 100.84 (+24.94) \\
\midrule
Romance  & French     & 13.90 & \textbf{11.03} (-2.87) & 13.29 & \textbf{10.80} (-2.49) 
                         & 52.25 & \textbf{18.96} (-33.29) & 35.88 & \textbf{26.09} (-9.79) \\
         & Galician   & 102.36 & \textbf{14.23} (-88.13) & 16.74  & \textbf{14.58} (-2.16) 
                         & 61.56 & \textbf{32.67} (-28.89) & \textbf{43.03} & 43.70 (+0.67) \\
         & Portuguese & 10.27 & \textbf{7.43} (-2.84)  &  8.74 & \textbf{7.73} (-1.01) 
                         & 25.05 & \textbf{12.29} (-12.76) & 18.31 & \textbf{14.09} (-4.22) \\
         & Romanian   & 47.85 & \textbf{18.03} (-29.82) & 17.02 & \textbf{14.89} (-2.13) 
                         & 93.31 & \textbf{26.79} (-66.52) & \textbf{30.04} & 29.20 (-0.84) \\
         & Spanish    &  8.34 & \textbf{6.66} (-1.68)  &  8.86 & \textbf{5.50} (-3.36) 
                         & 49.59 & \textbf{26.03} (-23.56) & \textbf{29.64} & 30.10 (+0.46) \\
\midrule
Slavic  & Belarusian & 100.82 & \textbf{30.13} (-70.69) & 25.51  & \textbf{24.13} (-1.38) 
                         & 183.00 & \textbf{62.35} (-120.65) & 60.21  & \textbf{51.05} (-9.16) \\
        & Polish     & 119.71 & \textbf{16.71} (-103.00) & 17.86  & \textbf{16.35} (-1.51) 
                         & 376.28 & \textbf{33.73} (-342.55) & 42.05  & \textbf{32.48} (-9.57) \\
        & Russian    & 108.17 & \textbf{12.46} (-95.71) & 14.62  & \textbf{13.11} (-1.51) 
                         & 158.95 & \textbf{42.50} (-116.45) & \textbf{42.77}  & 43.88 (+1.11) \\
        & Serbian    & 33.99  & \textbf{30.58} (-3.41)  & 30.31  & \textbf{27.91} (-2.40) 
                         & 63.18  & \textbf{33.15} (-30.03) & 34.50  & \textbf{31.69} (-2.81) \\
        & Slovenian  & 109.02 & \textbf{30.49} (-78.53) & 30.50  & \textbf{25.83} (-4.67) 
                         & 92.37  & \textbf{49.02} (-43.35) & 45.13  & \textbf{47.60} (+2.47) \\
\midrule
Turkic  & Azerbaijani & 104.03 & \textbf{47.22} (-56.81) & 27.51 & \textbf{26.23} (-1.28) 
                         & 145.09 & \textbf{79.00} (-66.09) & 143.74 & \textbf{85.73} (-58.01) \\
        & Turkish     & \textbf{21.55}  & 38.98 (+17.43) & 20.21 & \textbf{18.51} (-1.70) 
                         & 62.75  & \textbf{55.35} (-7.40)  & 68.74  & \textbf{57.49} (-11.25) \\
        & Kazakh      & 66.30  & \textbf{39.26} (-27.04) & \textbf{21.12} & 21.16 (+0.04) 
                         & \textbf{62.01}  & 69.48 (+7.47)  & \textbf{52.10}  & 64.78 (+12.68) \\
        & Kyrgyz      & \textbf{36.82}  & 53.47 (+16.65) & 30.82 & \textbf{30.33} (-0.49) 
                         & \textbf{68.69}  & 87.49 (+18.80) & 91.40  & \textbf{88.64} (-2.76) \\
\bottomrule
\end{tabular}
\end{threeparttable}
\end{table*}

\begin{table*}[h]
\centering
\begin{threeparttable}
\caption{Cross-domain WER$\downarrow$ (\%) per language using Gemma. 
$\Delta_f$ = Fam -- Lang. Best in \textbf{bold}}
\label{tab:rq2_crossdomain}
\begin{tabular}{ll|ll|ll}
\toprule
\multirow{2}{*}{\textbf{Family}} & \multirow{2}{*}{\textbf{Language}} 
& \multicolumn{2}{c|}{\textbf{CV $\rightarrow$ FL}} 
& \multicolumn{2}{c}{\textbf{FL $\rightarrow$ CV}} \\
\cmidrule(lr){3-4} \cmidrule(lr){5-6}
& & \textbf{Lang} & \textbf{Fam ($\Delta_f$)} 
  & \textbf{Lang} & \textbf{Fam ($\Delta_f$)} \\
\midrule

Afro-Asiatic & Amharic  & \textbf{119.70} & 154.38 (+34.68) & \textbf{91.23} & 95.59 (+4.36) \\
             & Arabic   & 93.60 & \textbf{78.01} (-15.59) & 116.22 & \textbf{76.51} (-39.71) \\
             & Hausa    & 202.45 & \textbf{180.67} (-21.78) & \textbf{65.79} & 80.19 (+14.40) \\
             & Hebrew   & \textbf{92.47} & 101.35 (+8.88) & \textbf{44.53} & 50.93 (+6.40) \\
             & Maltese  & 194.60 & \textbf{150.86} (-43.74) & 97.87 & \textbf{73.74} (-24.13) \\
\midrule

Baltic & Latvian    & \textbf{127.89} & 128.01 (+0.12) & 91.65 & \textbf{54.89} (-36.76) \\
       & Lithuanian & \textbf{98.04} & 122.45 (+24.41) & 42.49 & \textbf{40.31} (-2.18) \\
\midrule

Celtic & Irish & 210.63 & \textbf{172.54} (-38.09) & \textbf{109.72} & 131.11 (+21.39) \\
       & Welsh & \textbf{120.70} & 155.43 (+34.73) & 102.27 & \textbf{81.35} (-20.92) \\
\midrule

Dravidian & Malayalam & \textbf{104.26} & 113.49 (+9.23) & \textbf{50.56} & 74.78 (+24.22) \\
          & Tamil     & \textbf{89.59} & 117.97 (+28.38) & 120.75 & \textbf{67.94} (-52.81) \\
          & Telugu    & \textbf{120.07} & 164.27 (+44.20) & \textbf{73.37} & 111.97 (+38.60) \\
\midrule

Germanic & Danish   & 121.21 & \textbf{62.57} (-58.64) & 34.67 & \textbf{24.63} (-10.04) \\
         & Dutch    & 89.28  & \textbf{48.38} (-40.90) & 21.12 & \textbf{19.64} (-1.48) \\
         & English  & 65.07  & \textbf{28.39} (-36.68) & 23.59 & \textbf{17.30} (-6.29) \\
         & German   & 108.71 & \textbf{61.84} (-46.87) & 23.57 & \textbf{22.18} (-1.39) \\
         & Swedish  & 212.82 & \textbf{60.42} (-152.40) & 34.00 & \textbf{28.07} (-5.93) \\
\midrule

Indo-Iranian & Bengali & 119.81 & \textbf{103.87} (-15.94) & \textbf{40.16} & 41.89 (+1.73) \\
             & Hindi   & 110.46 & \textbf{88.96} (-21.50) & \textbf{49.34} & 59.05 (+9.71) \\
             & Persian & 214.23 & \textbf{186.80} (-27.43) & 142.85 & \textbf{119.56} (-23.29) \\
             & Punjabi & \textbf{78.34} & 139.35 (+61.01) & \textbf{37.81} & 56.84 (+19.03) \\
             & Urdu    & 94.79 & \textbf{89.74} (-5.05) & 152.09 & \textbf{134.70} (-17.39) \\
\midrule

Niger-Congo & Igbo    & \textbf{110.32} & 125.20 (+14.88) & \textbf{75.97} & 97.39 (+21.42) \\
            & Swahili & 199.47 & \textbf{187.92} (-11.55) & \textbf{56.27} & 94.87 (+38.60) \\
            & Yoruba  & \textbf{92.67} & 95.23 (+2.56) & \textbf{67.33} & 80.43 (+13.10) \\
\midrule

Romance & French     & 106.50 & \textbf{58.13} (-48.37) & 41.50 & \textbf{35.73} (-5.77) \\
        & Galician   & 86.53  & \textbf{81.75} (-4.78)  & \textbf{26.57} & 29.00 (+2.43) \\
        & Portuguese & 82.82  & \textbf{56.48} (-26.34) & 15.89 & \textbf{12.25} (-3.64) \\
        & Romanian   & 120.29 & \textbf{76.23} (-44.06) & 29.77 & \textbf{29.34} (-0.43) \\
        & Spanish    & 79.45  & \textbf{44.99} (-34.46) & 22.05 & \textbf{17.81} (-4.24) \\
\midrule

Slavic & Belarusian & \textbf{97.15} & 108.45 (+11.30) & \textbf{35.48} & 36.42 (+0.94) \\
       & Polish     & 105.60 & \textbf{63.44} (-42.16)  & 29.51 & \textbf{25.71} (-3.80) \\
       & Russian    & \textbf{90.24} & 98.83 (+8.59)    & 24.40 & \textbf{22.61} (-1.79) \\
       & Serbian    & 149.45 & \textbf{116.11} (-33.34) & 278.90 & \textbf{117.43} (-161.47) \\
       & Slovenian  & 113.94 & \textbf{80.06} (-33.88)  & 79.51 & \textbf{51.00} (-28.51) \\
\midrule

Turkic & Azerbaijani & \textbf{108.60} & 118.26 (+9.66) & 47.11 & \textbf{43.61} (-3.50) \\
       & Turkish     & 131.39 & \textbf{113.38} (-18.01) & 40.78 & \textbf{37.12} (-3.66) \\
       & Kazakh      & \textbf{108.74} & 146.46 (+37.72) & 65.02 & \textbf{61.34} (-3.68) \\
       & Kyrgyz      & \textbf{106.19} & 109.15 (+2.96)  & \textbf{64.54} & 66.95 (+2.41) \\
\bottomrule
\end{tabular}
\end{threeparttable}
\end{table*}

\begin{table*}
\centering
\caption{Detailed WER$\downarrow$ (\%) per language using Gemma across \textsc{LangConn}, \textsc{FamConn} and \textsc{UniConn}. Best in \textbf{bold}.}
\begin{tabular}{llccc}
\toprule
\multirow{2}{*}{\textbf{Family}} & \multirow{2}{*}{\textbf{Language}} 
& \multicolumn{3}{c}{\textbf{FLEURS}} \\
\cmidrule(lr){3-5}
& & \textbf{Lang} & \textbf{Fam} & \textbf{Uni} \\
\midrule

\multirow{5}{*}{Afro-Asiatic}
 & Amharic & \textbf{52.29} & 52.89 & 62.33 \\
 & Arabic  & 29.83 & 30.17 & \textbf{25.58} \\
 & Hausa   & \textbf{51.32} & 57.48 & 78.79 \\
 & Hebrew  & \textbf{42.90} & 46.59 & 53.02 \\
 & Maltese & \textbf{40.37} & 42.41 & 51.15 \\
\midrule

\multirow{2}{*}{Baltic}
 & Latvian    & 28.27 & \textbf{24.63} & 25.22 \\
 & Lithuanian & 34.56 & \textbf{30.18} & 34.16 \\
\midrule

\multirow{2}{*}{Celtic}
 & Irish  & 71.54 & \textbf{70.99} & 84.97 \\
 & Welsh  & 38.74 & \textbf{38.51} & 45.53 \\
\midrule

\multirow{3}{*}{Dravidian}
 & Malayalam & \textbf{21.85} & 22.95 & 33.57 \\
 & Tamil     & 32.24 & \textbf{31.92} & 40.70 \\
 & Telugu    & \textbf{34.34} & 35.16 & 50.77 \\
\midrule

\multirow{5}{*}{Germanic}
 & Danish   & 24.33 & \textbf{19.24} & 21.45 \\
 & Dutch    & 15.33 & \textbf{12.62} & 13.34 \\
 & English  &  8.59 & \textbf{6.79}  &  6.96 \\
 & German   & 17.95 & \textbf{13.34} & 13.51 \\
 & Swedish  & 20.61 & \textbf{16.56} & 18.83 \\
\midrule

\multirow{5}{*}{Indo-Iranian}
 & Bengali & 23.95 & \textbf{23.64} & 32.05 \\
 & Hindi   & 21.4 & \textbf{19.57} & 24.99 \\
 & Persian & \textbf{25.12} & 30.88 & 27.63 \\
 & Punjabi & 26.71 & \textbf{25.34} & 31.68 \\
 & Urdu    & 85.45 & \textbf{33.71} & 110.27 \\
\midrule

\multirow{3}{*}{Niger-Congo}
 & Igbo    & \textbf{55.25} & 56.69 & 80.69 \\
 & Swahili & \textbf{24.91} & 28.53 & 31.14 \\
 & Yoruba  & \textbf{45.51} & 46.63 & 55.14 \\
\midrule

\multirow{5}{*}{Romance}
 & French     & 13.29 & \textbf{10.80} & 11.55 \\
 & Galician   & 16.74 & \textbf{14.58} & 17.60 \\
 & Portuguese &  8.74 &  7.73          & \textbf{7.39} \\
 & Romanian   & 17.02 & \textbf{14.89} & 15.64 \\
 & Spanish    &  8.86 & \textbf{5.50}  &  6.25 \\
\midrule

\multirow{5}{*}{Slavic}
 & Belarusian & \textbf{25.51} & 24.13 & 29.03 \\
 & Polish     & \textbf{17.86} & 16.35 & 17.69 \\
 & Russian    & 14.62         & \textbf{13.11} & 13.39 \\
 & Serbian    & 30.31          & \textbf{27.91} & 30.61 \\
 & Slovenian  & 30.50          & \textbf{25.83} & 28.51 \\
\midrule

\multirow{4}{*}{Turkic}
 & Azerbaijani & 27.51 & \textbf{26.23} & 31.99 \\
 & Turkish     & 20.21 & \textbf{18.51} & 18.85 \\
 & Kazakh      & \textbf{21.12} & 21.16 & 26.23 \\
 & Kyrgyz      & 30.82 & \textbf{30.33} & 38.27 \\
\bottomrule
\end{tabular}
\label{tab:uni_detailed}
\end{table*}

\begin{table*}[!h]
\caption{Repeat rate and overlong rate of low-WER and high-WER cases. Repeat Rate:the proportion of predictions in which any word or phrase appears three or more times consecutively. Overlong Rate:the proportion of predictions where the output length exceeds 1.25 times the label length.}
\label{tab:extreme}
\resizebox{\linewidth}{!}{
\begin{tabular}{llccc}
\hline
\textbf{Language} & \textbf{Setting} & \multicolumn{1}{l}{\textbf{WER}} & \multicolumn{1}{l}{\textbf{Repeat Rate}} & \multicolumn{1}{l}{\textbf{Overlong Rate}} \\ \hline
English    & Fam\_FL\_Gemma       & 6.79   & 0.01 & 0.02 \\
Portuguese & Lang\_CV\_Salamandra & 15.72  & 0.01 & 0.01 \\
Hindi      & Fam\_CV\_Salamandra  & 35.35  & 0.02 & 0.06 \\
Polish     & Lang\_FL\_Salamandra & 119.71 & 0.94 & 0.45 \\
Welsh      & Fam\_CV\_Gemma       & 159.53 & 0.61 & 0.35 \\
Swahili    & Lang\_CV\_Salamandra & 482.6  & 0.87 & 0.64 \\ \hline
\end{tabular}}
\end{table*}

\end{document}